\documentclass[11pt,a4paper]{article}
\usepackage{times,latexsym}
\usepackage{url}
\usepackage[T1]{fontenc}

\usepackage[acceptedWithA]{tacl2021v1}

\usepackage{xspace,mfirstuc,tabulary}

\newif\iftaclinstructions
\taclinstructionsfalse 
\iftaclinstructions
\renewcommand{\confidential}{}
\renewcommand{\anonsubtext}{(No author info supplied here, for consistency with
TACL-submission anonymization requirements)}
\newcommand{\instr}
\fi

\iftaclpubformat 

\else

\fi

\usepackage{color}

\usepackage{amsmath}
\usepackage{booktabs}
\usepackage{graphicx}
\usepackage{multirow}
\usepackage{makecell}
\usepackage{algorithm}
\usepackage{inconsolata}
\usepackage{fontawesome}
\usepackage{algorithmicx}
\usepackage{algpseudocode}

\graphicspath{{img/},}

\title{Joint Speech Transcription and Translation: \\Pseudo-Labeling with Out-of-Distribution Data}

\author{%
Mozhdeh Gheini$^\diamond$\Thanks{Work done during an internship at Apple.} , Tatiana Likhomanenko$^\dagger$, Matthias Sperber$^\dagger$, Hendra Setiawan$^\dagger$ \\
$^\diamond$Information Sciences Institute,
University of Southern California \\
$^\dagger$Apple \\
\texttt{gheini@isi.edu, \{antares,sperber,hendra\}@apple.com} \\
}

\date{}

\begin{document}
\maketitle
\begin{abstract}
  Self-training has been shown to be helpful in addressing data scarcity for many domains, including vision, speech, and language. Specifically, self-training, or pseudo-labeling, labels unsupervised data and adds that to the training pool. In this work, we investigate and use pseudo-labeling for a recently proposed novel setup: joint transcription and translation of speech, which suffers from an absence of sufficient data resources. We show that under such data-deficient circumstances, the unlabeled data can significantly vary in domain from the supervised data, which results in pseudo-label quality degradation. We investigate two categories of remedies that require no additional supervision and target the domain mismatch: pseudo-label filtering and data augmentation. We show that pseudo-label analysis and processing as such results in additional gains on top of the vanilla pseudo-labeling setup resulting in total improvements of up to 0.6\% absolute WER and 2.2 BLEU points.
\end{abstract}

\section{Introduction}
\label{sec:intro}

Semi-supervised learning methods have been a cornerstone in addressing annotated data scarcity by taking advantage of and incorporating the relatively larger amounts of \textit{unlabeled}\footnote{We use descriptors ``(un)labeled'' and ``(un)supervised'' interchangeably throughout this draft.} data in the training process. Self-training is a relatively early instance of such methods \cite{1053799}. Conceptually, self-training is simple: first, a base model is trained using limited labeled data. The base model is then used to predict labels for the unlabeled data. The generated labels are termed ``\textit{pseudo-labels}'' to signify their predicted nature, as opposed to gold supervised data. Finally, the pseudo-labels are combined with the initial seed supervised data to train a new model, and this process is repeated until no further improvement in performance is observed.

Self-training, or pseudo-labeling interchangeably, has been shown to be effective to improve upon fully supervised baselines in low-resource settings for several sequence-to-sequence (seq2seq) tasks, such as machine translation (MT) \cite{zhang2018joint,DBLP:conf/iclr/HeGSR20,jiao2021self}, end-to-end speech recognition (ASR) \cite{xu20b_interspeech,park20d_interspeech,9054295,likhomanenko21b_interspeech}, and end-to-end speech translation (ST) \cite{pino20_interspeech}. In this work, we study pseudo-labeling for a recently proposed new setup, joint speech transcription and translation (STT)~\cite{anastasopoulos2018tied,sperber-etal-2020-consistent}: a setup that is of interest in use cases where both the transcript and translation of a speech signal are returned to the user. As we describe in detail later in \S\ref{subsec:stt}, the fully supervised data for modeling end-to-end joint transcription and translation is triples of form $(s, tc, tl)$ where $s$ is the speech signal, $tc$ is the transcript, and $tl$ is the translation. As that is especially costly to come by, STT also seems to have the potential to benefit from pseudo-labeling.

Our investigations show that while pseudo-labeling is indeed helpful, the quality of pseudo-labels that bring about the benefits is subpar. Upon inspecting the supervised and unsupervised sets, that proves to be not surprising: with limited amounts of supervised data, it is likely that the supervised and unsupervised sets differ in domain, impacting the quality of pseudo-labels. Specifically, in our case, we identify two causes leading to domain mismatch with out-of-distribution unlabeled data: difference between the sequence length ranges and vocabulary sets of the supervised and unsupervised sets. In this work, we ask \textit{if} we can specifically counteract the domain mismatch to reach a set of pseudo-labels of higher quality, and \textit{if} that higher quality, in turn, translates into a better overall performance of pseudo-labeling.

First, we propose pseudo-label filtering, which is often a part of pseudo-labeling algorithm~\cite{9054295,park20d_interspeech,zhang2021flexmatch,likhomanenko21b_interspeech,zhang2022censer}. However, while filtering is usually based on the model prediction scores, we rely on data-centric criteria~\cite{likhomanenko21b_interspeech} that directly target the identified domain mismatch aspects. 
Second, we augment the supervised data by concatenating randomly-picked samples to create new ones and adding them to the supervised set. These two are essentially different in nature: while filtering increases the overall quality by removing samples with pseudo-labels that are likely to be faulty, augmentation does so by extending the supervised set and generating better labels in the first place. Our results confirm that indeed this distinction in nature gets reflected in different ways filtering and augmentation improve the performance of pseudo-labeling.

The outline of this paper is as follows. We provide some background in \S\ref{sec:backgrnd} and detail the experimental setup in \S\ref{sec:exp}. Then, in \S\ref{sec:res}, we report and discuss the results from vanilla pseudo-labeling, the observation of domain mismatch, and the gains brought about by filtering and augmentation.

Our \textbf{contributions} are: 1) We specifically focus on pseudo-labeling in the face of domain mismatch between the supervised and unsupervised sets; 2) We investigate the mitigation of the effect of domain mismatch through two approaches: pseudo-label filtering and augmentation by concatenation and demonstrate how they improve pseudo-labeling in different ways. These approaches can be repurposed wherever pseudo-labeling is considered as a solution; 3) We apply pseudo-labeling modified with those approaches specifically to a novel setup, joint speech transcription and translation, and report gains on top of the vanilla pseudo-labeling for STT.

\section{Background}
\label{sec:backgrnd}

Our work studies a pseudo-labeling solution for end-to-end joint speech transcription and translation. In this section, we provide the background for these two components involved in the study, namely \emph{speech transcription and translation} and \emph{pseudo-labeling}.

\begin{algorithm*}
\caption{Pseudo-labeling}\label{alg:1}
\begin{algorithmic}[1]
\Require $L = \{x_i, y_i\}$ and $U = \{x_j\}$
\State Train base model $M$ on $L$
\While{The desired number of rounds or convergence has not been reached}
    \State Generate the pseudo-labeled set: $P = \{{x_j, M(x_j)} \mid x_j \in U\}$
    \State Obtain $M^+$ by fine-tuning $M$ on $L \cup P$
    \State Replace $M$ with $M^+$
\EndWhile
\State \Return $M$
\end{algorithmic}
\end{algorithm*}

\subsection{Speech Transcription and Translation}
\label{subsec:stt}

Our task of speech transcription and translation (STT) is closely related to script recognition (ASR) and speech translation (ST). ASR is the task of generating the text equivalent to an audio speech signal. Meanwhile, ST aims to generate the text equivalent to the signal in a target language other than the language of the speaker. In contrast, STT generates both the transcript and the translation \textit{jointly} in an end-to-end fashion. STT is particularly appealing in cases where both the transcript and translation are to be displayed to the user.

Formally, STT can be modeled as follows: given a speech signal ($s$), the model generates the transcript ($tc$) and translation ($tl$) concatenated together in the output as one single sequence: $s \rightarrow tc \_ tl$ \cite{sperber-etal-2020-consistent}. This formulation is simple to implement as it casts STT as an instance of the well-known seq2seq modeling and results in a single end-to-end model to be stored on device. Furthermore, as reported by \newcite{sperber-etal-2020-consistent}, this formulation results in a reasonably consistent transcripts and translations as the coupled inference ensures that translations are conditioned on transcripts. In our experiments, we adopt this STT formulation.

However, the major challenge that such modeling presents is insufficient data resources: three-way parallel samples of form $(s, tc, tl)$ are expensive to annotate. Annotation would require multilingual annotators and would be time-consuming. To alleviate this limitation, we study how pseudo-labeling can be employed effectively to combat data scarcity in this setting. We provide a background on pseudo-labeling in the next section.



\subsection{Pseudo-labeling}
\label{subsec:pl}
Pseudo-labeling, which is also often referred to as self-training in the literature, addresses the data insufficiency issue by taking advantage of much larger amounts of unsupervised data. More precisely, assume a labeled set $L = \{x_i, y_i\}$ and an unlabeled set $U = \{x_j\}$, where $|U| \geq |L|$,  are available (note that in the case of STT, $y_i$ is actually a tuple consisting of the transcript and the translation: $y_i = (tc_i, tl_i)$). Pseudo-labeling starts with training an initial model $M$ in a supervised manner using $L$. Then, using $M$, it generates pseudo-labels (predictions) for $U$. It then incorporates the pseudo-labels to create a new model $M^+$, which hopefully supersedes $M$ in performance. $M^+$ can then replace $M$ to repeat this process for as many rounds as desired, or until no further gains are observed. Although conceptually simple, several key decisions need to be made before pseudo-labeling can be applied:

\begin{itemize}
    \item \emph{How should $M^+$ be created?} $M^+$ can be trained from scratch (e.g., as done by \newcite{park20d_interspeech}) or alternatively obtained by continuously fine-tuning $M$ (e.g., as done by \newcite{xu20b_interspeech}) using the labeled set combined with the pseudo-labeled set. As we later report in \S\ref{sec:res}, in our preliminary experiments, fine-tuning consistently outperforms training from scratch. Hence, we opt for fine-tuning in our experiments.
    \item \emph{Should pseudo-labeling be applied to supervised sets?} For the pseudo-labeling stage, we consider and experiment with labeling the supervised set in addition to the unsupervised set and monitor for any potential improvements. Similar to the previous item, as we later show in \S\ref{sec:res}, using the pseudo-labels for the supervised set does not prove to be beneficial in our preliminary experiments. Therefore, we generate predictions only for the unlabeled set.
    \item \emph{In what way should the pseudo-labels be used to update existing models?} For instance, \newcite{DBLP:conf/iclr/HeGSR20}, at each round, first train a model from scratch on the pseudo-labeled set, and then fine-tune it on the labeled set to obtain the final model for that round. Alternatively, \newcite{xu20b_interspeech} combine the two sets and use a hyper-parameter to have control over the relative weight of the labeled portion against the pseudo-labeled portion. To keep our setup simple, we opt for combining the sets and treating them equally.
\end{itemize}

With the key factors outlined above, Algorithm~\ref{alg:1} shows how we carry out vanilla pseudo-labeling for our experiments. All results we report in \S\ref{subsec:vpl} follow this algorithm.

\section{Experimental Setup}
\label{sec:exp}

We describe the supervised set and the unsupervised set for our experiments  in \S\ref{subsec:data} and our model architecture in \S\ref{subsec:model}.

\subsection{Data}
\label{subsec:data}
In this work, we use two publicly available multilingual speech translation datasets which, thanks to the nature of their creation, also include transcripts: CoVoST V2 \cite{wang2020covost} and MuST-C \cite{CATTONI2021101155}. CoVoST V2 is created by amending the validated audio clips and transcripts from the Common Voice crowd-sourced ASR corpus \cite{ardila-etal-2020-common} with professional translations. It covers translations from English into 15 languages and from 21 languages into English. MuST-C is created by automatically aligning the audio segments from TED talks to corresponding manual transcripts and translations (available from the TED website), which are also aligned. It covers translations from English into 14 languages.

We conduct our experiments across two language pairs : English--German (En--De) and English--Chinese (En--Zh), which are available in both CoVoST and MuST-C. In all our experiments, we designate CoVoST as the supervised set, and MuST-C as the unsupervised set. Note that this means our objective is to reach the best performance possible on the CoVoST evaluation set. While we also have the gold transcripts and translations (labels in the STT problem) for MuST-C, we do not use them and practically treat MuST-C as an unlabeled set. We only use MuST-C gold labels for analysis and pseudo-label quality assessment. We provide the statistics of our data in Table~\ref{tab:datastat}.

\begin{table}[t]
    \centering
    \begin{tabular}{lcccc}
        \toprule
         & \multicolumn{2}{c}{CoVoST} & \multicolumn{2}{c}{MuST-C} \\
         & Train & Eval & Train & Eval \\
        \midrule
        En--De & 233k & 15.5k & 251k & 1.4k \\
        En--Zh & 233k & 15.5k & 359k & 1.3k  \\
        \bottomrule
    \end{tabular}
    \caption{Amount of data available (number of sentences), per language pair and corpus.}
    \label{tab:datastat}
\end{table}

\subsection{Model}
\label{subsec:model}
Our model is an end-to-end Transformer \cite{NIPS2017_3f5ee243} with a hidden dimension of 1024, and five and three layers of encoder and decoder respectively (following \newcite{sperber-etal-2020-consistent}). For the input, on the encoder side, we first use wav2vec 2.0 \textsc{Base} \cite{NEURIPS2020_92d1e1eb} as provided by \texttt{Hugging Face Transformers} \cite{wolf-etal-2020-transformers} (specifically, \texttt{facebook/wav2vec2-base-960h}) to extract speech representations before feeding them into the Transformer encoder. On the output side, as described in \S\ref{subsec:stt}, the decoder generates one sequence consisting of the transcript and the translation concatenated together.

In terms of input prepossessing, we remove instances where speech is either shorter than 0.5 seconds or longer than 15 seconds, or either the transcript or the translation is longer than 50 words. After that, we use \texttt{SentencePiece} \cite{kudo-richardson-2018-sentencepiece} for subword tokenization. We use a vocabulary size of 1020 and 8188 in the case of En--De and En--Zh, respectively. The transcription and translation vocabulary is shared in both cases.

The objective function during optimization is a weighted sum of the CTC loss \cite{10.1145/1143844.1143891} on the encoder side, and the cross-entropy loss on the decoder side. Both during training a base model and fine-tuning an existing checkpoint on the union of the labeled set and the pseudo-labeled set, we use Adam optimizer \cite{DBLP:journals/corr/KingmaB14} with peak learning rate of 0.0005 after 500 warmup steps, coupled with inverse square root learning rate scheduling. We train for a total of 100 epochs.

For both language pairs, we use the dev sets provided by the corpora as the held-out evaluation set. We remove diacritics and punctuation before scoring and report our performance in terms of WER of transcripts and BLEU of translations using beam size of five with \textsc{Sacre}BLEU.\footnote{Hash: case.lc+numrefs.1+smooth.4.0+tok.\{13a,zh\} for \{En--De,En--Zh\}.}

Our implementation is built upon \texttt{PyTorch} \cite{Paszke_PyTorch_An_Imperative_2019}, \texttt{xnmt} \cite{neubig-etal-2018-xnmt}, and \texttt{Lightning} \cite{Falcon_PyTorch_Lightning_2019}.


\begin{table*}[t]
    \centering
    \scalebox{0.85}{\begin{tabular}{clrrrrrrrrrrr}
        \toprule
         & & \multicolumn{5}{c}{En--De} & \multicolumn{6}{c}{En--Zh} \\
        \cmidrule(r){3-7}
        \cmidrule(l){8-13}
         & & Base Model & R1 & R2 & R3 & Bound & Base Model & R1 & R2 & R3 & R4 & Bound \\
        \cmidrule(r){3-7}
        \cmidrule(l){8-13}
        \multirow{2}{*}{\faSearch CoVoST} & WER $\downarrow$ & 15.4 & 15.4 & 15.0 & 15.0 & 14.4 & 14.8 & 14.6 & 14.8 & 14.7 & 14.6 & 13.7 \\
         & BLEU $\uparrow$ & 22.8 & 23.8 & 24.5 & 24.5 & 25.5 & 28.7 & 29.4 & 30.0 & 30.5 & 30.7 & 31.9 \\
        \cmidrule(r){1-7}
        \cmidrule(l){8-13}
        \multirow{2}{*}{MuST-C} & WER $\downarrow$ & 45.1 & 45.2 & 29.7 & 28.4 & 9.6 & 47.9 & 46.2 & 43.8 & 42.8 & 37.2 & 8.9 \\
         & BLEU $\uparrow$ & 7.3 & 9.1 & 9.7 & 9.6 & 22.4 & 9.1 & 9.9 & 9.6 & 9.0 & 8.3 & 18.9 \\
        \bottomrule
    \end{tabular}}
    \caption{Vanilla pseudo-labeling results over each round up to saturation. CoVoST, our supervised set, is distinguished with \faSearch  symbol to signify it is intended to improve performance on it.}
    \label{tab:baseres}
\end{table*}

\section{Results and Discussion}
\label{sec:res}

We present our results in this section in the following order: \S\ref{subsec:vpl} establishes vanilla pseudo-labeling performance, which leads to our analysis of the domain mismatch between the supervised and unsupervised sets. \S\ref{subsec:filt} and \S\ref{subsec:aug} then describe the two categories of remedies we devise to mitigate the effect of domain discrepancies on pseudo-labeling.

\begin{figure}[t]
    \centering
    \includegraphics[width=0.9\columnwidth]{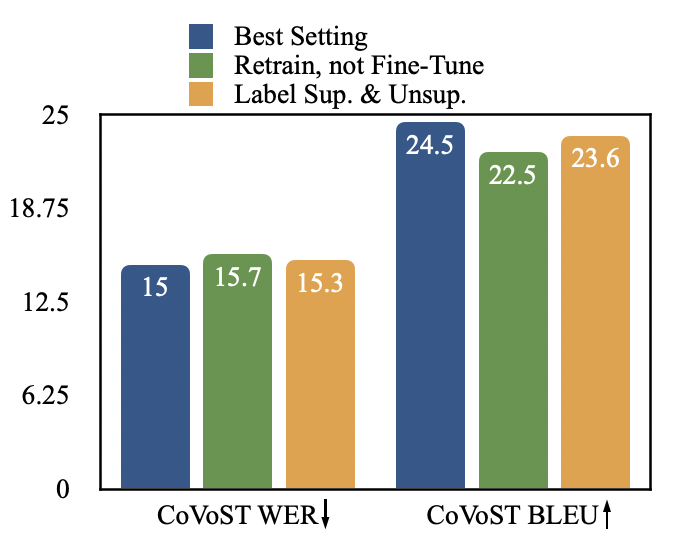}
    \caption{Performance of different PL settings on the supervised set: CoVoST. The best setting fine-tunes the checkpoint from the last round on the supervised set and the pseudo-labels for the unsupervised set.}
    \label{fig:pl_setting}
\end{figure}

As mentioned in \S\ref{subsec:pl}, this is all using the best setting we were able to establish during our pilot experiments: at each pseudo-labeling round, we 1) label only the unsupervised data, and 2) fine-tune the existing checkpoint on the combination of supervised and pseudo-labeled data. We conduct our pilot experiments on En--De. We were able to confirm that the aforementioned setting consistently beats the rest over several rounds of pseudo-labeling. Figure~\ref{fig:pl_setting} illustrates the lead of the best setting over others in the last round of our experiments. The same pattern holds across all rounds.

\subsection{Vanilla Pseudo-Labeling}
\label{subsec:vpl}
In Table~\ref{tab:baseres}, we include the results of vanilla pseudo-labeling, as in Algorithm~\ref{alg:1}, with no modifications. We report WER and BLEU for En--De and En--Zh across both corpora. To reiterate, CoVoST (distinguished by the magnifying glass symbol \faSearch) is our designated supervised set, and hence, what we are trying to boost performance on. MuST-C scores, on the other hand, are reported for the sake of analysis; the metrics are to assess the quality of the pseudo-labels.

We report the performance of the initial model (the fully supervised baseline, Model $M$ on line 1 of the Algorithm~\ref{alg:1}) in the ``Base Model'' column. Scores from each pseudo-labeling round, thereafter, appear on the corresponding ``R'' column. To have an upper bound of what is possible with the collective data that we have, we run an experiment training a single model using both corpora in a supervised manner. Those numbers are provided in the ``Bound'' column. Note that this is the only case for which MuST-C gold labels are used, solely to obtain an upper bound performance: what would be possible if we predicted labels perfectly for the unsupervised set? Is pseudo-labeling helping us close the gap between the base model and that upper bound for \faSearch CoVoST?

First and foremost, in confirmation with the literature, vanilla pseudo-labeling is effective. On \faSearch CoVoST, it is able to improve the base model by 0.4\% absolute WER and 1.7 BLEU points on En--De, and 0.2\% absolute WER and 2.0 BLEU points on En--Zh. However, with a closer look at the quality of pseudo-labels at each round (i.e., MuST-C scores), it is evident that the generated labels are far from ideal quality.  Specifically, even if we train plain machine translation systems on \faSearch CoVoST transcripts and translations (and take the audio out of the picture), the En--De system scores 12.4 BLEU on MuST-C En--De, and the En--Zh system scores 9.6 BLEU on MuST-C En--Zh.

\begin{table*}[t]
    \centering
    \scalebox{0.9}{\begin{tabular}{lrrrrrrrr}
        \toprule
         & \multicolumn{4}{c}{En--De} & \multicolumn{4}{c}{En--Zh} \\
        \cmidrule(r){2-5}
        \cmidrule(l){6-9}
         & \multicolumn{2}{c}{\faSearch CoVoST} & \multicolumn{2}{c}{MuST-C} & \multicolumn{2}{c}{\faSearch CoVoST} & \multicolumn{2}{c}{MuST-C} \\
         & WER $\downarrow$ & BLEU $\uparrow$ & \multirow{2}{*}{WER $\downarrow$} & \multirow{2}{*}{BLEU $\uparrow$} & WER $\downarrow$ & BLEU $\uparrow$ & \multirow{2}{*}{WER $\downarrow$} & \multirow{2}{*}{BLEU $\uparrow$} \\
        Bound & 14.4 & 25.5 & & & 13.7 & 31.9 & & \\
        \cmidrule(r){1-5}
        \cmidrule(l){6-9}
        Vanilla PL & 15.4/\textbf{15.0} & 23.8/24.5 & 45.2/28.4 & 9.1/9.7 & 14.6/14.6 & 29.4/30.7 & 46.2/37.2 & 9.9/9.9 \\
        \cmidrule(r){1-5}
        \cmidrule(l){6-9}
        Ratio to Gold & 15.3/15.0 & 24.1/24.7 & 22.8/15.8 & 9.6/10.4 & 14.5/14.2 & 29.5/30.5 & 23.2/17.4 & 10.0/10.2 \\
        Ratio KDE & \textbf{15.1}/\textbf{15.0} & 24.2/24.5 & 30.5/27.1 & 9.4/10.1 & \textbf{14.3}/\textbf{14.2} & 29.8/30.7 & 30.8/21.7 & 10.8/10.8 \\
        LASER & 15.2/\textbf{15.0} & 24.1/24.5 & 34.7/27.6 & 9.6/10.0 & 14.6/14.3 & 29.4/30.6 & 40.8/20.3 & 10.7/11.2 \\
        \cmidrule(r){1-5}
        \cmidrule(l){6-9}
        Augmentation & 15.3/15.3 & \textbf{24.9}/\textbf{24.9} & 33.8/22.2 & 11.5/11.8 & 14.6/14.3 & \textbf{30.1}/\textbf{30.9} & 48.7/25.4 & 11.9/11.9 \\
        \bottomrule
    \end{tabular}}
    \caption{Improved results using remedies recommended. Each cell includes the performance obtained from the first round and the best performance obtained using the corresponding method (R1/Best). We also include bounds from Table~\ref{tab:baseres} for \faSearch CoVoST for comparison. We use bold font to mark the best performance on \faSearch CoVoST.}
    \label{tab:betterres}
\end{table*}

\begin{figure}[t]
    \centering
    \includegraphics[width=\columnwidth]{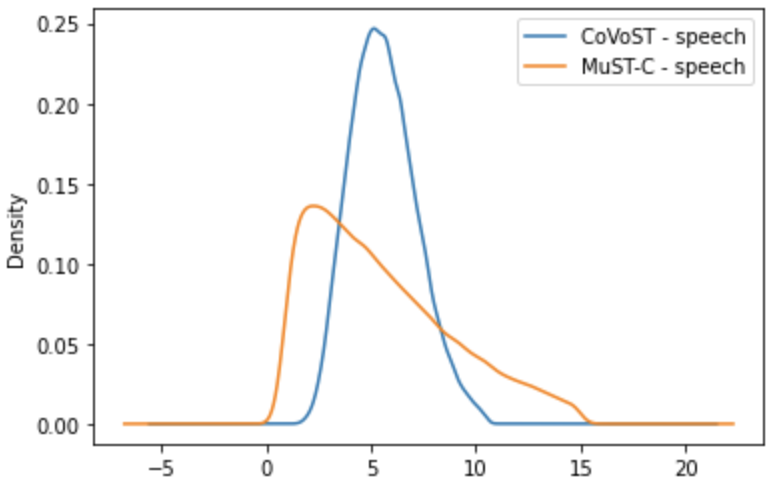}
    \vspace{-0.8cm}
    \caption{The probability density function of input speech lengths estimated using kernel density estimation. MuST-C speech signals are longer in duration.}
    \label{fig:length_kde}
\end{figure}

Our investigation into the reasons as to why that is the case points to two root causes that indicate \faSearch CoVoST and MuST-C are significantly different in \textit{domain} in the following aspects:

\begin{itemize}
    \item Length mismatch between corpora: As shown in Figure~\ref{fig:length_kde}, MuST-C speech sequences are generally longer, which also results in longer transcripts and translations.
    \item Vocabulary mismatch between corpora: Additionally, we were able to identify discrepancies between the vocabulary of words between the two corpora. For instance, on the English side, MuST-C and CoVoST each have roughly 64k and 121k unique types, respectively. Of those, only 38k types are in common, with CoVoST having more probability mass on rare (tail-end of the Zipfian distribution) vocabulary types.
\end{itemize}

Following this observation, we next demonstrate that it is possible to counteract the domain mismatch and enhance the quality of labels to boost the effectiveness of pseudo-labeling.

\subsection{Direction \#1: Data-Centric Filtering}
\label{subsec:filt}

Per \S\ref{subsec:pl}, in vanilla pseudo-labeling, we use all the generated labels to update the model. Alternatively, pseudo-labels can be filtered to remove predictions of less quality. Recent works \cite{park20d_interspeech} rely on confidence scores from the model to filter the pseudo-labels, which require careful and proper normalization. \newcite{9054295} use a combination of heuristic-based and confidence-based filtering. In our case, similar to \newcite{likhomanenko21b_interspeech}, we propose and only rely on data-centric metrics to specifically target domain-mismatch and select a subset of pseudo-labels to use in the next round: transcript length to speech length ratio and transcript and translation LASER embeddings cosine similarity.

\subsubsection{Length Ratio Distribution}
A sign of flawed inference and faulty output in seq2seq models has been known to be looping \cite{ChorowskiJ17}: the model generates the same n-gram repeatedly. We were also able to identify looping occurring frequently in the pseudo-labels and resulting in long transcripts. While the supposed lengths of the correct transcripts are unknown, the length of the input audio can be used as an indicator: heuristically, the shorter the input audio, the shorter the transcript.

\begin{figure*}
    \centering
    \includegraphics[width=\textwidth]{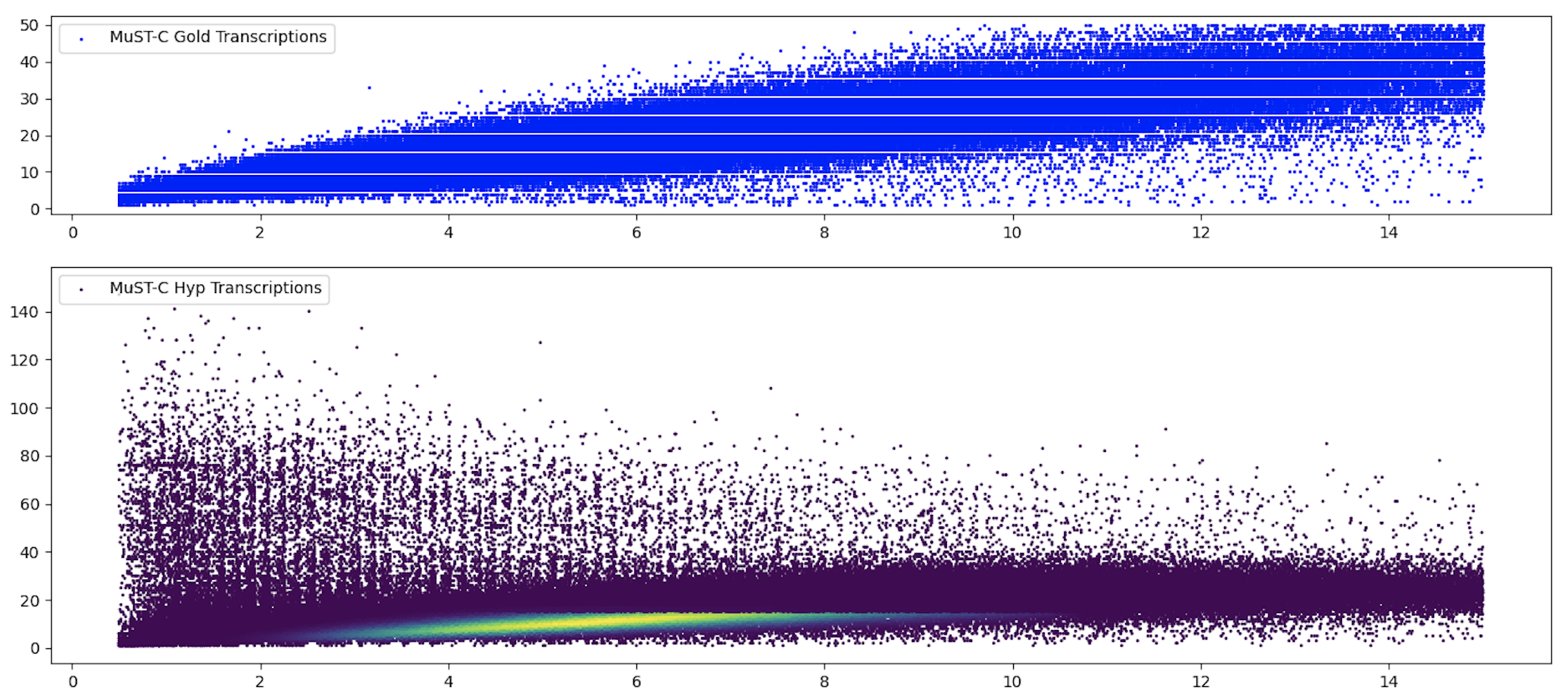}
    \vspace{-1cm}
    \caption{Plots of transcript lengths against input audio lengths (gold transcripts above and generated transcripts during pseudo-labeling at the bottom). Datapoints in the below plot are color-coded based on their PDF value as estimated by KDE, with lighter colors indicating higher values. The most probable mass forms a pattern similar to that of gold transcription. Therefore PDF values can be effective for filtering pseudo-labels of less quality.}
    \label{fig:length_ratio_kde}
\end{figure*}

To take advantage of this signal with no supervision overhead, we estimate the probability density function (PDF) (using kernel density estimation (KDE)) of the joint probability distribution over the input speech lengths and predicted transcripts lengths. At each pseudo-labeling round then, we only keep the top 90\% of the most probable transcripts. Figure~\ref{fig:length_ratio_kde} visualizes the effect of such filtering. Instances that have the highest PDF values, have a similar ratio of transcript length to speech length to that of gold transcripts. Hence, this can be a useful metric that needs no supervision.

To gauge the maximum potential effectiveness of length ratio-based filtering, we also conduct experiments with filtering based on the ratio of the generated transcript length to the \textit{gold} transcript length, where we only keep those with the length within 0.9$\times$ and 1.1$\times$ the length of the corresponding gold transcript. Note that this only has discussion purposes, as it uses supervision in the form of access to the length of the gold transcripts.

Table~\ref{tab:betterres} (rows ``Ratio to Gold'' and ``Ratio KDE'') shows how our length ratio-based filtering methods compare against plain vanilla pseudo-labeling. For each method, we run the same number of rounds as we did for vanilla pseudo-labeling in Table~\ref{tab:baseres}. We report the performance of the first round and the best round (first round/best round in table cells) of each method. Results from each separate round are comprehensively provided in Appendix~\ref{app:extres}.

On \faSearch CoVoST, ``Ratio KDE'' speeds up gains relative to vanilla pseudo-labeling despite incorporating fewer labels (only 90\%): 15.1 vs. 15.4 WER and 24.2 vs. 23.8 BLEU at the first round in the case of En--De. The same pattern holds for En--Zh. Looking at the scores on MuST-C, it is evident that moderating the quality of pseudo-labels in this way, does indeed translate into better pseudo-labels for future rounds and improved performance on the supervised set. Also, ``Ratio to Gold'', benefiting from a form of supervision, expectedly results in better quality on the unsupervised set. However, on the supervised set, it performs similarly to ``Ratio KDE'', demonstrating that ``Ratio KDE'' is effective enough at removing detrimental pseudo-labels.

While ``Ratio KDE'' performs clearly better at earlier rounds, it saturates at the same performance as vanilla pseudo-labeling, which uses all the labels (with being better only in the case of En--Zh WER by 0.4\% absolute WER). So it is especially beneficial when available resources can only cover a small number of pseudo-labeling rounds.

\subsubsection{LASER Score}
Our second filtering method relies on the relationship between the generated translations and transcripts (in contrast to the previous method, which relied on the relationship between the generated transcripts and speech signals). For this, we use LASER \cite{artetxe-schwenk-2019-massively}, a multilingual sentence encoder, to embed the generated transcripts and translations in a multilingual space to rank pairs based on cosine similarity and hold onto only the top 90\%. Given that LASER lies at the center of this, the quality of representations of different languages in its multilingual space can affect the degree of usefulness.

Per Table~\ref{tab:betterres}, row ``LASER'', in our experiments LASER-based filtering improves performance on the unsupervised set (and hence, the quality of the pseudo-labels) all across the board. Those improvements translate into better performance on the supervised set in the case of En--De. Importantly, the improvement pattern is similar to that of length ratio-based filtering: more gains at earlier rounds, saturating at the same performance as the vanilla pseudo-labeling. So similarly, LASER-based filtering can be useful when fewer runs can be run.

\subsection{Direction \#2: Data Augmentation}
\label{subsec:aug}

Our previous filtering methods remove pseudo-labels so that the remaining subset has a higher quality. However, if we can generate better labels, to begin with, we can discard none and retain all the labels. Here, to improve the quality of the labels generated by the base model at no extra supervision cost, we use data augmentation by concatenation to directly target the reported length mismatch between corpora in \S\ref{subsec:vpl}. To do so, we create an augmented set from our supervised set by randomly selecting a pair of samples and constructing a new sample by concatenating the speech signals as the input and concatenating corresponding transcripts and translations as output. In our experiments, we build a set of 20k augmented samples as such using the original \faSearch CoVoST data. After training the base model, before generating pseudo-labels, we first further fine-tune the base model on the union of the original supervised set and the augmented set. We then proceed as in vanilla pseudo-labeling with the union of the original data and the augmented set as our supervised training set.

As shown in Table~\ref{tab:betterres}, row ``Augmentation'', we can see that although no generated labels are thrown away, the quality of pseudo-labels is indeed increased in the subsequent round. This is especially pronounced in the case of translations.

With retaining all pseudo-labels, not only does bootstrapping the supervised set using concatenation expedite the gains from pseudo-labeling, but it is also the most effective in terms of the final performance before saturation by improving the score in three cases: it improves the performance of vanilla pseudo-labeling on \faSearch CoVoST by 0.4 and 0.2 BLEU points on En--De and En--Zh, respectively, and by 0.3\% absolute WER on En--Zh. Therefore, it further closes the gap between pseudo-labeling and the upper bounds.

This ends our discussion of how domain mismatch can be addressed. We find filtering methods, which discard labels, to be only effective when due to any resource limitation, only a few, one or two rounds of pseudo-labeling can be run. This finding also echoes insights from \newcite{pmlr-v162-bansal22b} that studies data scaling laws for MT and shows while filtering may benefit computational efficiency, more unfiltered data can replace filtered data. As an alternative to filtering, we show that improving the quality of all generated labels through augmentation so that all can be kept, is the most effective, especially when as many rounds as needed can be run to reach saturation.

\section{Related Work}

The two paradigms often considered in low-resource data scenarios are self-training and pretraining. Self-training and pseudo-labeling have long been studied for a variety of seq2seq tasks \cite{DBLP:conf/iclr/HeGSR20,xu20b_interspeech,park20d_interspeech,9054295,DBLP:conf/interspeech/ChenWW20,likhomanenko21b_interspeech,pino20_interspeech}. Regarding the relationship between pretraining and self-training, \newcite{9414641} and \newcite{wang21r_interspeech} show that self-training and unsupervised pretraining are complimentary and can be combined to boost performance on speech recognition and speech translation, respectively. In the case of supervised pretraining, however, \newcite{NEURIPS2020_27e9661e} show in the vision domain that as the size of the labeled data available grows, self-training remains helpful, whereas the benefits of supervised pretraining start to diminish.

In the application of self-training to the unvisited setup of joint speech transcription and translation \cite{sperber-etal-2020-consistent}, we focus on domain mismatch, a matter which can get overlooked when gains from vanilla pseudo-labeling are observed. For our solutions, we study pseudo-label filtering and augmentation by concatenation. In contrast to conventional filtering which relies on normalized model confidence scores \cite{park20d_interspeech,9054295}, we define and use data-centric factors that directly target the domain differences that we observe.

Concatenation as an effective augmentation method has been studied in the context of machine translation \cite{Agrawal2018ContextualHI,kondo-etal-2021-sentence,nguyen-etal-2021-data,https://doi.org/10.48550/arxiv.2210.05096} and speech-to-text \cite{DBLP:journals/corr/abs-2210-15398}. In our case, we use it to expose our base model to sequences of higher length to improve the quality of generated pseudo-labels.

\section{Conclusion}

We studied pseudo-labeling for joint speech transcription and translation. We show that while vanilla pseudo-labeling is helpful, there are additional improvements to be gained from addressing the low quality of generated pseudo-labels due to the domain mismatch between the supervised and unsupervised sets.

We find that our proposed solutions help in two different ways, as they are in distinct nature: pseudo-label filtering, which discards low-quality labels, is mostly helpful by expediting gains in earlier rounds, especially for transcriptions. Augmentation by concatenation, on the other hand, does not discard any of the labels. As a result, it is able to maintain an edge over vanilla pseudo-labeling in the late rounds as well.


\section*{Acknowledgements}
We would like to thank Qin Gao, Amittai Axelrod, Boliang Zhang, Barry Theobald, David Grangier, Jiatao Gu and the rest of machine translation and machine learning research teammates for fruitful discussions and constructive feedback on the manuscript.

\bibliography{custom,acl_anthology}
\bibliographystyle{acl_natbib}

\appendix

\section{Extended Results}
\label{app:extres}

\begin{table*}[t]
    \centering
    \scalebox{0.9}{\begin{tabular}{lrrrr}
    \toprule
     & \multicolumn{4}{c}{En--De} \\
    \midrule
     & \multicolumn{2}{c}{\faSearch CoVoST} & \multicolumn{2}{c}{MuST-C} \\
     & WER $\downarrow$ & BLEU $\uparrow$ & \multirow{2}{*}{WER $\downarrow$} & \multirow{2}{*}{BLEU $\uparrow$} \\
    Bound & 14.4 & 25.5 & & \\
    \midrule
    Base Model & 15.4 & 22.8 & 45.1 & 7.3 \\
    \midrule
    \multirow{3}{*}{Vanilla PL} & 15.4 & 23.8 & 45.2 & 9.1 \\
     & 15.0 & 24.5 & 29.7 & 9.7 \\
     & 15.0 & 24.5 & 28.4 & 9.6 \\
    \midrule
    \multirow{3}{*}{Ratio to Gold} & 15.3 & 24.1 & 22.8 & 9.6 \\
     & 15.0 & 24.5 & 18.5 & 10.2 \\
     & 15.1 & 24.7 & 15.8 & 10.4 \\
    \midrule
    \multirow{3}{*}{Ratio KDE} & 15.1 & 24.2 & 30.5 & 9.4 \\
     & 15.0 & 24.5 & 27.7 & 9.8 \\
     & 15.4 & 24.4 & 27.1 & 10.1 \\
    \midrule
    \multirow{3}{*}{LASER} & 15.2 & 24.1 & 34.7 & 9.6 \\
     & 15.0 & 24.5 & 29.1 & 9.9 \\
     & 15.3 & 24.5 & 27.6 & 10.0 \\
    \midrule
    \multirow{2}{*}{Augmentation} & 15.3 & 24.9 & 33.8 & 11.5 \\
     & 15.3 & 24.9 & 22.2 & 11.8 \\
    \bottomrule
    \end{tabular}}
    \caption{Extended results on En--De. All run until saturation. Each row represents one round of pseudo-labeling with the respective method.}
    \label{tab:extres_de}
\end{table*}

\begin{table*}[t]
    \centering
    \scalebox{0.9}{\begin{tabular}{lrrrr}
    \toprule
     & \multicolumn{4}{c}{En--Zh} \\
    \midrule
     & \multicolumn{2}{c}{\faSearch CoVoST} & \multicolumn{2}{c}{MuST-C} \\
     & WER $\downarrow$ & BLEU $\uparrow$ & \multirow{2}{*}{WER $\downarrow$} & \multirow{2}{*}{BLEU $\uparrow$} \\
    Bound & 13.7 & 31.9 & & \\
    \midrule
    Base Model & 14.8 & 28.7 & 47.9 & 9.1 \\
    \midrule
    \multirow{3}{*}{Vanilla PL} & 14.6 & 29.4 & 46.2 & 9.9 \\
     & 14.8 & 30.0 & 43.8 & 9.6 \\
     & 14.7 & 30.5 & 42.8 & 9.0 \\
     & 14.6 & 30.7 & 37.2 & 8.3 \\
    \midrule
    \multirow{3}{*}{Ratio to Gold} & 14.5 & 29.5 & 23.2 & 10.0 \\
     & 14.3 & 30.5 & 18.7 & 10.2 \\
     & 14.4 & 30.5 & 17.9 & 9.7 \\
     & 14.2 & 30.5 & 17.4 & 9.9 \\
    \midrule
    \multirow{3}{*}{Ratio KDE} & 14.3 & 29.8 & 30.8 & 10.8 \\
     & 14.3 & 30.2 & 22.0 & 10.8 \\
     & 14.2 & 30.4 & 21.7 & 10.8 \\
     & 14.2 & 30.7 & 21.7 & 10.4 \\
    \midrule
    \multirow{3}{*}{LASER} & 14.6 & 29.4 & 40.8 & 10.7 \\
     & 14.4 & 30.4 & 27.5 & 10.4 \\
     & 14.4 & 30.5 & 24.4 & 10.4 \\
     & 14.3 & 30.6 & 20.3 & 11.2 \\
    \midrule
    \multirow{3}{*}{Augmentation} & 14.6 & 30.1 & 48.7 & 11.9 \\
     & 14.5 & 30.5 & 35.7 & 11.0 \\
     & 14.5 & 30.9 & 26.3 & 11.5 \\
     & 14.3 & 30.9 & 25.4 & 11.3 \\
    \bottomrule
    \end{tabular}}
    \caption{Extended results on En--Zh. All run until saturation. Each row represents one round of pseudo-labeling with the respective method.}
    \label{tab:extres_zh}
\end{table*}

\end{document}